\documentclass[biblatex, english]{lni}
\usepackage{booktabs}

\usepackage{tabularx}

\usepackage[]{blindtext}

\begin{document}
%%% Mehrere Autoren werden durch \and voneinander getrennt.
%%% Die Fußnote enthält die Adresse sowie eine E-Mail-Adresse.
%%% Das optionale Argument (sofern angegeben) wird für die Kopfzeile verwendet.
\title[Privacy-Preserving Text Preprocessing]{Current State in Privacy-Preserving Text Preprocessing for Domain-Agnostic NLP}
%%%\subtitle{Untertitel / Subtitle} % falls benötigt
\author[1]{Abhirup Sinha\textsuperscript{\textdagger}}{abhirup@mail.upb.de}{0000-0002-6927-5526}
\author[1]{Pritilata Saha\textsuperscript{\textdagger}}{psaha@mail.upb.de}{0000-0002-7776-1620}
\author[2]{Tithi Saha\textsuperscript{\textdagger}}{tithi.saha2020@vitstudent.ac.in}{0009-0006-8887-5806}%
\affil[1]{Paderborn University\\Department of Computer Science\\Warburger Straße 100\\33098 Paderborn\\Germany}
\affil[2]{Vellore Institute of Technology\\School of Computer Science and Engineering\\Vellore, Tamil Nadu 632014\\India}
\maketitle
\def\thefootnote{\textsuperscript{\textdagger}}\footnotetext{The authors contributed equally to this work}\def\thefootnote{\arabic{footnote}}

\begin{abstract}
Privacy is a fundamental human right. Data privacy is protected by different regulations, such as GDPR. However, modern large language models require a huge amount of data to learn linguistic variations, and the data often contains private information. Research has shown that it is possible to extract private information from such language models. Thus, anonymizing such private and sensitive information is of utmost importance. While complete anonymization may not be possible, a number of different pre-processing approaches exist for masking or pseudonymizing private information in textual data. This report focuses on a few of such approaches for domain-agnostic NLP tasks.
\end{abstract}

%TODO

\section{Introduction}
\label{sec:intro}
% brief outlook on privacy rights
Many text sources contain personal information that can be traced back to an individual. Thus, the usage of such text data without explicit consent violates an individual's privacy. As privacy is a fundamental human right, the use of text data containing personal information falls under the General Data Protection Regulation (GDPR) \citep{lison-etal-2021-anonymisation}. 

% danger in text privacy with LLM advancements
Training of Large Language Models (LLMs), a new advancement in NLP, requires a lot of data from different sources to capture linguistic variations. However, recent works show that training data can be extracted from these LLMs, which gives out personal information about people like names, email addresses, phone numbers, etc. \citep{carlini2021extracting, nasr2023scalable}. According to GDPR, the presence of such personal information in training data requires prior informed consent. Leakage of personal information as part of training data can be considered a data breach, and it is a punishable offence, according to GDPR. 

% What is Text Anonymization
However, as prior consent is often difficult to obtain, GDPR permits anonymizing personal data \citep{yermilov-etal-2023-privacy} so that it can not be used to re-identify a data subject. It involves the removal of personally identifiable information (PII) from text data. In this process, PII, which can directly (e.g., name, passport number, etc.) or indirectly (e.g., gender, nationality, etc.) identify an individual, is removed, masked or substituted \citep{lison-etal-2021-anonymisation}. For successful anonymization, individuals should not be re-identified even after combining several indirect substituted identifiers. 

Anonymization of textual data for clinical or legal use cases is very crucial, and many studies have been done on these domains \citep{johnson2020deidentification, csanyi2021challenges, garat2022automatic}. However, it is very difficult to transfer existing anonymization models trained on one domain (e.g., clinical) to new domains (e.g., legal) \citep{hartman2020customization}. Not many studies treated the text anonymization problem in a domain-independent way. Thus, this short paper focuses on identifying the currently existing domain-independent methods for anonymization during text pre-processing. This paper also states their limitations and future scope of work in this direction.

\section{Approaches for Text Anonymization during Preprocessing}
\label{sec:methods}
Some domain-agnostic approaches for anonymization during text preprocessing have already been proposed. We initially did keyword search on ACL Anthology, then further refined it via forward and backward literature search. Among the approaches, \citet{papadopoulou-etal-2022-bootstrapping, olstad-etal-2023-generation} proposed anonymization with replacements from Wikidata ontology. \citet{yermilov-etal-2023-privacy} also performed an interesting case study on the pseudonymization techniques. These existing state-of-the-art approaches are discussed in the subsequent sections.

\subsection{Pseudonymization}
\label{subsec:pseudo}
Pseudonymization is an approach to anonymizing PII, which is a little different than its hypernym de-identification in a broad NLP sense. For de-identification or destructive anonymization, masked texts have the PII-related information replaced by some generic identifiers or labels. However in pseudonymized texts, PII-related information is replaced by other substitutes, so that the text still looks realistic \citep{eder-etal-2022-beste}. Table \ref{tab:text-comparison} provides a comparison between masked de-identified text and pseudonymized text against a reference text sample. 

\begin{table}[htbp]
\centering
\begin{tabularx}{\columnwidth}{>{\hsize=0.2\hsize}X|X}
\toprule
\textbf{Reference text}     & John, an engineer at Microsoft in California, collaborates with Mary.                  \\ \midrule
\textbf{Masked text}     & PERSON\_1, an engineer at ORGANIZATION\_1 in LOCATION\_1, collaborates with PERSON\_2. \\ \midrule
\textbf{Pseudonymized text} & Mike, an engineer at Apple in New York, collaborates with Steve. \\ \bottomrule
\end{tabularx}
\caption{An example comparison between a masked text sample and a pseudonymized text sample for a given reference text with personally identifiable information (PII).}
\label{tab:text-comparison}
\end{table}

\citet{yermilov-etal-2023-privacy} investigated the effectiveness of three different pseudonymization techniques with NLP models. They pseudonymized three categories of named entities: PERSON, LOC (Location), and ORG (Organization). With pseudonymized training datasets, the authors evaluated the downstream performance of the models on text classification and summarization tasks. The three pseudonymization techniques- NER-based, Seq2Seq, and LLM-based, are discussed briefly in the following sections. 

\paragraph{NER-based Pseudonymization:}
In NER-based pseudonymization, the authors used named entity recognition (NER) models to detect PII-containing text spans. The detected text spans were replaced with similar types of named entities from Wikidata Knowledge Graph. \citet{yermilov-etal-2023-privacy} generated a list of replacement candidates first, and then one random replacement candidate was chosen, following some predefined constraints. 

\paragraph{Seq2Seq Pseudonymization;}
In Seq2Seq pseudonymization, the task of pseudonymization was treated as a sequence-to-sequence (Seq2Seq) task. The authors used BART \citep{lewis-etal-2020-bart}. This BART model was finetuned on a corpus of pseudonymized texts, generated by NER-based pseudonymization techniques. 

\paragraph{LLM-based Pseudonymization:}
For LLM-based pseudonymization, \citet{yermilov-etal-2023-privacy} used two pre-trained LLMs- GPT-3 \citep{brown2020language}, and ChatGPT (GPT-3.5). These two LLMs were used sequentially. GPT-3 was used to extract named entities from texts. ChatGPT was used to perform pseudonymization on the extracted named entities. ChatGPT was preferred over GPT-3 in the pseudonymization task because of its better qualitative performance \citep{yermilov-etal-2023-privacy}. 

% TODO: Shorten
\subsection{Ontology and Rule-based Approaches}
\label{subsec:onto}
\citet{papadopoulou-etal-2022-bootstrapping} treated the problem of text anonymization as a token-level sequence classification task. Their approach involved identifying PII-related text sequences by constructing an inverted index from any knowledge graph. For experiments, they considered a subset of the Wikidata KG, consisting of entities such as names, nicknames, professions, etc. \citet{papadopoulou-etal-2022-bootstrapping} used RoBERTa \citep{liu2019roberta} for entity detection on a short dataset of Wikipedia biographies. They considered the following categories for entity masking- PERSON, LOC, ORG, DEM (Demographic), DATETIME, QUANTITY and MISC (Miscellaneous). \textit{k}-anonymity (introduced by \citet{samarati2001protecting}) was used to ensure each personally identifiable (PII-related) entity is indistinguishable from at least $k-1$ other entities of similar category. In case of violation, their algorithm had two choices- selecting the term with the shortest posting in the inverted index (Greedy Selection) or  selecting a term randomly, irrespective of its posting in the inverted index (Random Selection). 

\citet{olstad-etal-2023-generation} expanded upon the work of \citet{papadopoulou-etal-2022-bootstrapping}. At first, various text spans containing PII were detected using sequence labelling models (detailed by \citet{lison-etal-2021-anonymisation}), and the corresponding type was determined. They considered 8 different categories of PII identifiers, following \citet{pilan-etal-2022-text}. For entities of type PERSON, QUANTITY and DATETIME, heuristic rules were used because they can not usually be a part of a privacy-focused ontology \citep{olstad-etal-2023-generation}. Entities of type DEM, LOC, ORG and MISC (e.g., events, nationalities, works of art, etc.) were replaced by suitable generalizations found in the ontology. If no match was found in the local ontology, a Wikidata query was done. Entities of type CODE (e.g., credit card numbers, SSN numbers) were masked as `***'. This work further enriched the dataset from \citet{papadopoulou-etal-2022-bootstrapping}.

% TODO: Merge Limitations and Future Works, also shorten and add a few lines on metrics. Keep only the major limitations
\section{Current Limitations and Future Works}
\label{sec:limitations}
%only focus on English
A major limitation of the discussed approaches is that the experimentations were only performed on the English datasets. However, many languages (e.g., German, Mandarin, etc.) have other tokenization schemes that significantly differ from English. Privacy-sensitive texts can also be found in other languages as well. Thus, more works are needed to be done in languages other than English. \citet{yermilov-etal-2023-privacy} also pointed out future work in this direction.

A major problem in this field of text anonymization is the limited availability of annotated corpora. Annotation works for text anonymization are more costly and time-consuming than regular annotation works \citep{papadopoulou-etal-2022-bootstrapping}. Production of silver corpora with automated annotation can alleviate this problem. Works of \citet{papadopoulou-etal-2022-bootstrapping, olstad-etal-2023-generation} are targeted towards this direction, but more works are needed. Also, as we mentioned in Section \ref{sec:intro}, quite a few works on text anonymization exist in the clinical or legal NLP domain. However, recent training data extraction techniques \citep{carlini2021extracting, ishihara-2023-training, zhang-etal-2023-ethicist, nasr2023scalable} from LLMs pose a serious threat to common person's privacy. More work is needed for developing general-purpose domain-agnostic approaches, models, and corpora. 

%pseudonymization-specific
For the pseudonymization studies, \citet{yermilov-etal-2023-privacy} considered a limited subset of named entity types. They only focused on entities of type PERSON, Locations (LOC), and Organizations (ORG). However, PII can also be of other named entity types, as shown by \citet{pilan-etal-2022-text}. Studying the performance of pseudonymization approaches on other PII-entity types could give us a whole picture. 

%ontology-specific
For the ontology-driven PII-masking approach of \citet{olstad-etal-2023-generation} and \citet{papadopoulou-etal-2022-bootstrapping}, 
results show an over-masking tendency, resulting in low data utility \citep{papadopoulou-etal-2022-bootstrapping}. Moreover, an ontology-driven approach could also suffer from ambiguities arising from entity-linking in ontology \citep{olstad-etal-2023-generation, papadopoulou-etal-2022-bootstrapping}. Also, it would be interesting to see how such multiple approaches can be combined, and is it better than the already discussed approaches.

Also, the already discussed different anonymization approaches used different NLP metrics to report their results. There was no apparent way to compare between the approaches. Also, the standard NLP metrics, e.g., precision, recall, F-Score, have various shortcomings in anonymization tasks, which were pointed out by \citet{pilan-etal-2022-text}. Future works in this direction should uniformly adopt suitable metrics (e.g., Entity-level Recall on Direct Identifiers, Entity-level Recall on Quasi-Identifiers and Token-level Weighted Precision on both Direct and Quasi-Identifiers, as proposed by \citet{pilan-etal-2022-text}). This also makes any comparison among various approaches easier. 

%report limitations
Lastly, this paper only covered the few existing domain-independent approaches to anonymization during text preprocessing. We provided a high-level overview of those few approaches. But, we did not perform any experimentation. Many previous works on text anonymization focused on clinical NLP \citep{lison-etal-2021-anonymisation, johnson2020deidentification, hartman2020customization}. There are also some recent works that focus on text anonymization in legal NLP \citep{csanyi2021challenges, glaser2021anonymization, garat2022automatic}. So, more domain-specific approaches exist, which are out of scope for this paper.  

\section{Conclusion}
In this short paper, we focused on text anonymization techniques that can be used during data preprocessing. Text anonymization is a way of protecting an individual's privacy in textual documents. The focus of anonymization is not only on protecting privacy but also on preserving the utility of such documents. Although anonymization has been widely used in clinical or legal NLP, a few anonymization approaches exist for domain-independent NLP. Pseudonymization or Ontology-driven approaches work directly on data during text preprocessing. Ontology-driven approaches can also be used to construct corpora for text anonymization tasks. Finally, more work needs to be done to adapt text anonymization approaches in languages other than English.  

%% \bibliography{lni-paper-example-de.tex} ist hier nicht erlaubt: biblatex erwartet dies bei der Preambel
%% Starten Sie "biber paper", um eine Biliographie zu erzeugen.
\printbibliography

\end{document}